\pdfoutput=1
\documentclass[11pt,a4paper]{article}
\makeatletter
\providecommand{\@LN}[2]{}
\makeatother
\usepackage[hyperref]{acl2021}
\aclfinalcopy %
\def\aclpaperid{2898} %
\usepackage{times}
\usepackage{latexsym}
\usepackage{nicematrix}
\usepackage{inconsolata}

\usepackage[T1]{fontenc}

\usepackage[utf8]{inputenc}

\usepackage{CJKutf8}
\usepackage{microtype}

\usepackage{tikz}
\usepackage{amsmath}
\usepackage{amssymb}
\usepackage{amsthm}
\usepackage{booktabs}
\usepackage{enumitem}
\usepackage{multirow}
\usepackage[capitalize]{cleveref}
\usepackage{dsfont}
\usepackage{stmaryrd}
\usepackage{lipsum}
\usepackage{xspace}
\usepackage{pmboxdraw}

\definecolor{cb-blue}       {RGB}{ 0, 109, 219}
\definecolor{cb-burgundy}   {RGB}{146,   0,   0}
\definecolor{cb-green-lime} {RGB}{138, 226,  52}
\definecolor{cb-yellow}     {RGB}{253, 216, 53}

\usepackage[textsize=footnotesize]{todonotes}

\newcommand{\bigcell}[2]{\begin{tabular}{@{}#1@{}}#2\end{tabular}}

\newcommand{\stderr}[1]{\scriptsize $\pm #1$}

\newcommand{\mybarheight}{2mm}
\newcommand{\myboxwidth}{8mm}
\newcommand{\mybar}[2]{\framebox[\myboxwidth][l]{\rule{#1mm}{\mybarheight}}}
\usepackage{adjustbox}
\usepackage{array}
\usepackage{float}
\newcolumntype{R}[2]{%
    >{\adjustbox{angle=#1,lap=\width-(#2)}\bgroup}%
    l%
    <{\egroup}%
}

\newcommand*\REDC{\tikz[baseline=(char.base)]{
        \node[draw=red, fill=red, text=white, shape=circle, minimum size=4mm, inner sep=0pt] (char)
        {\rule[-3pt]{0pt}{\dimexpr2ex+2pt}r};}\xspace}
\newcommand*\BLUEC{\tikz[baseline=(char.base)]{
        \node[draw=cb-blue, fill=cb-blue, text=white, shape=circle, minimum size=4mm, inner sep=0pt] (char)
        {\rule[-3pt]{0pt}{\dimexpr2ex+2pt}b};}\xspace}
\newcommand*\YELC{\tikz[baseline=(char.base)]{
        \node[draw=cb-yellow, fill=cb-yellow, text=black, shape=circle, minimum size=4mm, inner sep=0pt] (char)
        {\rule[-3pt]{0pt}{\dimexpr2ex+2pt}y};}\xspace}
\newcommand*\GREENC{\tikz[baseline=(char.base)]{
        \node[draw=cb-green-lime, fill=cb-green-lime, text=black, shape=circle, minimum size=4mm, inner sep=0pt] (char)
        {\rule[-3pt]{0pt}{\dimexpr2ex+2pt}g};}\xspace}
        
\newcommand\vocab{\mathcal{V}}

\newcommand\boxRight{\textSFii\pmboxdrawuni{2574}}

\newcommand\boxSpace{\hspace{1.5em}}

\title{Lexicon Learning for Few-Shot Neural Sequence Modeling}

\author{Ekin Aky\"urek ~~~~ Jacob Andreas \\
  Massachusetts Institute of Technology \\
  \texttt{\{akyurek,jda\}@mit.edu}
  }

\begin{document}
\maketitle
\begin{abstract}
Sequence-to-sequence transduction is the core problem in language processing
  applications as diverse as semantic parsing, machine translation, and
  instruction following. The neural network models that provide the dominant
  solution to these problems are brittle, especially in low-resource settings:
  they fail to generalize correctly or systematically from small datasets. Past
  work has shown that many failures of systematic generalization arise from
  neural models' inability to disentangle \emph{lexical} phenomena from
  \emph{syntactic} ones. To address this, we augment neural decoders with a
  \emph{lexical translation mechanism} that generalizes existing copy mechanisms
  to incorporate learned, decontextualized, token-level translation rules. We
  describe how to initialize this mechanism using a variety of lexicon learning
  algorithms, and show that it improves systematic generalization on a diverse
  set of sequence modeling tasks drawn from cognitive science, formal semantics, and machine translation.\footnote{Our code is released under \url{https://github.com/ekinakyurek/lexical}}
\end{abstract}

\section{Introduction}
\begin{figure}[t]
\centering
\resizebox{\columnwidth}{!}{%
  \includegraphics[clip,trim=0 4.7in 6in 1.2in]{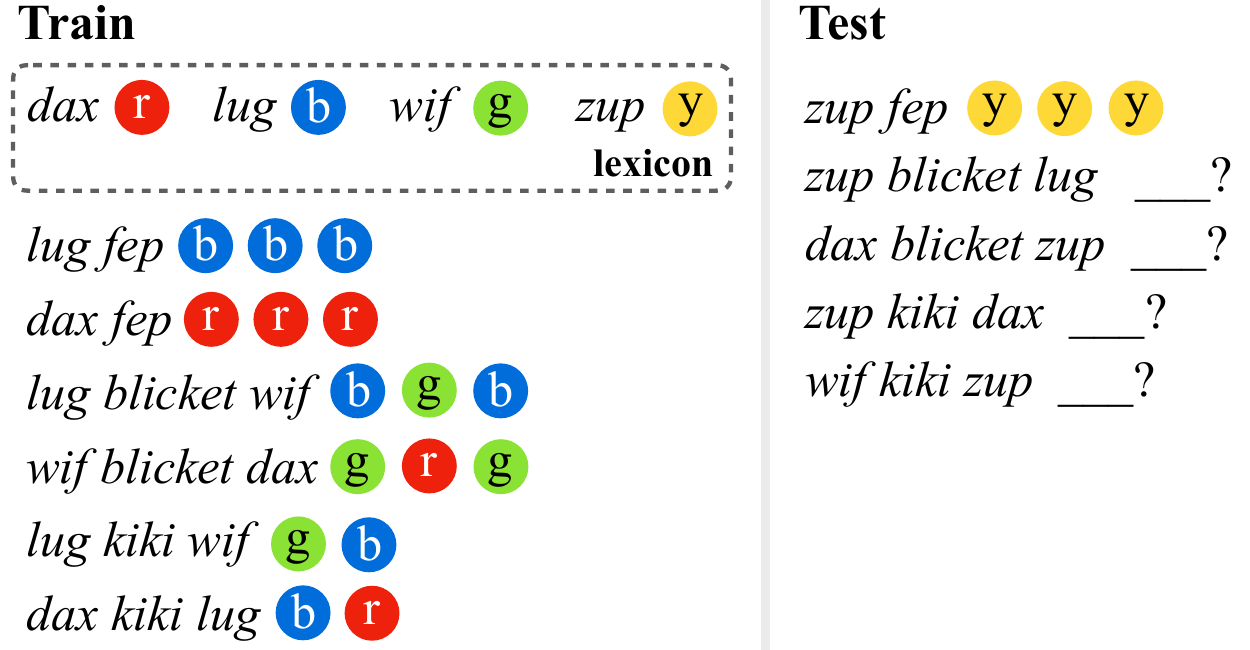}
}
\caption{A fragment of the Colors dataset from \citet{lake2019human}, a simple sequence-to-sequence translation task. The output vocabulary is only the colored circles \REDC, \GREENC, \BLUEC,  \YELC. %
Humans can reliably fill in the missing test labels on the basis of a small
  training set, but standard neural models cannot. This paper describes a neural
  sequence model that obtains improved generalization via a learned
  \emph{lexicon} of token translation rules.}
\label{tab:Colors}
\end{figure}

Humans exhibit a set of structured and remarkably consistent inductive biases when learning from language data. For example, in both natural language acquisition and toy language-learning problems like the one depicted in \cref{tab:Colors}, human learners exhibit a preference for systematic and compositional interpretation rules (\citealt{guasti2017language}, Chapter 4; \citealt{lake2019human}). These inductive biases in turn support behaviors like one-shot learning of new concepts \cite{carey1978acquiring}.
But in natural language processing, recent work  has found that state-of-the-art
neural models, while highly effective at in-domain prediction, fail to
generalize in human-like ways when faced with rare phenomena and small datasets
\cite{lake2018generalization}, posing a fundamental challenge for NLP tools in the low-data regime.

Pause for a moment to fill in the missing labels in \cref{tab:Colors}. While doing so, which training examples did you pay the most attention to? How many times did you find yourself saying \emph{means} or \emph{maps to}?
Explicit representations of lexical items and their meanings play a key role diverse models of syntax and semantics \citep{joshi1997tree, pollard1994head, bresnan2015lexical}.
But one of the main findings in existing work on generalization in \emph{neural} models is that they fail to cleanly separate \textbf{lexical} phenomena from \textbf{syntactic} ones \cite{lake2018generalization}. Given a dataset like the one depicted in \cref{tab:Colors}, models conflate (lexical) information about the correspondence between \emph{zup} and \YELC with the (syntactic) fact that \YELC appears only in a sequence of length 1 at training time. Longer input sequences containing the word \emph{zup} in new syntactic contexts cause models to output tokens only seen in longer sequences (\cref{sec:experiments}).

In this paper, we describe a parameterization for sequence decoders that facilitates (but does not enforce) the learning of context-independent word meanings. 
Specifically, we augment decoder output layers with a \textbf{lexical translation
mechanism} which generalizes neural copy mechanisms \citep[e.g.][]{see2017get}
and enables models to generate token-level translations purely attentionally.
While the lexical translation mechanism is quite general, we focus here on its
ability to improve few-shot learning in sequence-to-sequence models. On a suite
of challenging tests of few-shot semantic parsing and instruction following,
our model exhibits strong generalization, achieving the highest reported results for neural sequence models on datasets as diverse as COGS (\citealt{kim2020cogs}, with 24155 training examples) and Colors (\citealt{lake2019human}, with 14). Our approach also generalizes to real-world tests of few-shot learning, improving BLEU  scores \cite{papineni2002bleu} by 1.2 on a low-resource English--Chinese machine translation task (2.2 on test sentences requiring one-shot word learning). 

In an additional set of experiments, we explore effective procedures for
initializing the lexical translation mechanism using lexicon learning algorithms
derived from information theory, statistical machine translation, and Bayesian
cognitive modeling. We find that both mutual-information- and alignment- based
lexicon initializers perform well across tasks.
Surprisingly, however, we show that both approaches can be matched or
outperformed by a \emph{rule-based} initializer that identifies high-precision
word-level token translation pairs.   
We then explore joint learning of the lexicon and decoder, but find (again surprisingly) that this gives only marginal improvements over a fixed initialization of the lexicon. 

In summary, this work:
\begin{itemize}[]
    \item Introduces a new, lexicon-based output mechanism for neural encoder--decoder models.
    \item Investigates and improves upon lexicon learning algorithms for
    initialising this mechanism.
    \item Uses it to solve challenging tests of generalization in instruction following, semantic parsing and machine translation.
\end{itemize}

A great deal of past work has suggested that neural models come equipped with an inductive bias that makes them fundamentally ill-suited to human-like generalization about language data, especially in the low-data regime \citep[e.g.][]{fodor1988connectionism,marcus2018deep}.
Our results suggest that the situation is more complicated:
by offloading
the easier lexicon learning problem to simpler models, neural sequence models
are actually quite effective at modeling (and generalizing about) about syntax
in synthetic tests of generalization and real translation tasks.

\begin{CJK*}{UTF8}{bsmi}
\begin{table*}[t]
\centering
\resizebox{\textwidth}{!}{%
\begin{tabular}{@{}lll@{}}
\toprule
\multicolumn{1}{c}{\bf Inputs}  & \multicolumn{1}{c}{\bf Outputs}  & \multicolumn{1}{c}{\bf Lexicon Entries} \\ \hline %
\bigcell{l}{\it A crocodile blessed William . \\ \it William needed to walk . } & \bigcell{l}{\tt crocodile(x\_1) AND bless.agent (x\_2, x\_1) AND bless.theme (x\_2, William) \\ \tt need.agent (x\_1 , William) AND need.xcomp(x\_1, x\_3) AND walk.agent (x\_3, William)} & \bigcell{l}{\textit{blessed} $\mapsto$ \texttt{bless} \\ \textit{needed} $\mapsto$ \texttt{need} \\ \textit{William} $\mapsto$ \texttt{William}} \\ \midrule %
\bigcell{l}{\it Many moons orbit around Saturn \\ \it Earth is a planet .	} & \bigcell{l}{許多 \, 衛星 \, 繞著 \, 土星  \, 運行 . \\ 地球 \, 是 \, 一個  \, 行星 . } & \bigcell{l}{\textit{saturn} $\mapsto$ 土星 \\ \textit{earth} $\mapsto$ 地球 \\ \textit{moon} $\mapsto$ 衛星} \\ \midrule %
\textit{\bigcell{l}{walk around left \\ turn right \\ turn left \\ jump \\ jump opposite right after look left}} & \texttt{\bigcell{l}{LTURN IWALK LTURN IWALK LTURN IWALK LTURN IWALK \\ RTURN \\ LTURN \\ IJUMP \\ LTURN ILOOK RTRUN IJUMP RTURN IJUMP}}   &
\bigcell{l}{\textit{walk} $\mapsto$ \texttt{IWALK} \\ \textit{jump} $\mapsto$ \texttt{IJUMP} \\ \textit{right} $\mapsto$ \texttt{RTURN} \\ \textit{left} $\mapsto$ \texttt{LTURN} \\ \textit{look} $\mapsto$ \texttt{ILOOK}} \\ %
\bottomrule
\end{tabular}%

}
\caption{We present example (input,output) pairs from COGS, English-to-Chinese machine translation and SCAN datasets. We also present some of the lexicon entries which can be learned by proposed lexicon learning methods and that are  helpful to make generalizations required in each of the datasets.}
\label{tab:datasets}
\end{table*}
\end{CJK*}

\section{Related Work}\label{sec:related}
\paragraph{Systematic generalization in neural sequence models}

The desired inductive biases noted above are usually grouped together as ``systematicity'' but in fact involve a variety of phenomena: one-shot learning of new concepts and composition rules \cite{lake2018generalization}, zero-shot interpretation of novel words from context cues \cite{gandhi2020mutual}, and interpretation of known concepts in novel syntactic configurations \cite{keysers2019measuring, kim2020cogs}. What they share is a common expectation that learners should associate specific production or transformation rules with specific input tokens (or phrases), and generalize to use of these tokens in new contexts.

Recent years have seen tremendous amount of modeling work aimed at encouraging these generalizations in neural models, primarily by equipping them with symbolic scaffolding in the form of program synthesis engines \citep{nye2020learning}, stack machines \citep{grefenstette2015learning, liu2020compositional}, or symbolic data transformation rules \citep{gordon2019permutation,geca}. A parallel line of work has investigated the role of continuous representations in systematic generalization, proposing improved methods for pretraining \citep{furrer2020compositional} and procedures for removing irrelevant contextual information from word representations \citep{arthur2016incorporating,   russin2019compositional,thrush2020compositional}.
The latter two approaches proceed from similar intuition to ours, aiming to disentangle word meanings from syntax in encoder representations via alternative attention mechanisms and adversarial training. Our approach instead focuses on providing an explicit lexicon to the decoder; as discussed below, this appears to be considerably more effective.

\paragraph{Copying and lexicon learning}

In neural encoder--decoder models,
the clearest example of benefits from special treatment of word-level production rules is the \emph{copy mechanism}. A great deal of past work has found that neural models benefit from learning a structural copy operation that selects output tokens directly from the input sequence without requiring token identity to be carried through all neural computation in the encoder and the decoder. These mechanisms are described in detail in \cref{sec:approach-lex}, and are widely used in models for language generation, summarization and semantic parsing. Our work generalizes these models to structural operations on the input that replace copying with general context-independent token-level translation.

As will be discussed, the core of our approach is a (non-contextual) lexicon that maps individual input tokens to individual output tokens. Learning lexicons like this is of interest in a number of communities in NLP and language science more broadly. A pair of representative approaches \citep{brown1993mathematics,Frank2007ABF} will be discussed in detail below; other work on lexicon learning for semantics and translation includes \citet{liang2009learning,Johnson2007NonparametricBM,Haghighi2008LearningBL} among numerous others. 

Finally, and closest to the modeling contribution in this work, several previous
papers have proposed alternative generalized copy mechanisms for tasks other
than semantic lexicon learning. Concurrent work by \citet{Prabhu2020MakingAP}
introduces a similar approach for grapheme-to-phoneme translation (with a fixed
functional lexicon rather than a trainable parameter matrix), and
\citet{Nguyen2018ImprovingLC} and \citet{gu2019pointer} describe less
expressive mechanisms that cannot smoothly interpolate between 
lexical translation and ordinary decoding at the token level.
\citet{pham2018towards} incorporate lexicon entries by rewriting \emph{input}
sequences prior to ordinary sequence-to-sequence translation.
\citet{akyurek2020learning} describe a model in which a copy mechanism is
combined with a retrieval-based generative model; like the present work, that
model effectively disentangles syntactic and lexical information by using
training examples as implicit representations of lexical correspondences. 

We generalize and extend this previous work in a number of ways, providing a new
parameterization of attentive token-level translation and a detailed study of
initialization and learning. But perhaps the most important contribution of this work is the observation that many of the hard problems studied as ``compositional generalization'' have direct analogues in more conventional NLP problems, especially machine translation. Research on systematicity and generalization would benefit from closer attention to the ingredients of effective translation at scale.

\section{Sequence-to-Sequence Models With Lexical Translation Mechanisms}
\label{sec:approach-lex}

This paper focuses on sequence-to-sequence \textbf{language understanding} problems like the ones depicted in \cref{tab:datasets}, in which the goal is to map from a natural language \textbf{input} $x = [x_1, x_2, \ldots, x_n]$ to a structured \textbf{output} $y = [y_1, y_2, \ldots, y_m]$---a logical form, action sequence, or translation. We assume input tokens $x_i$ are drawn from a \textbf{input vocabulary} $\vocab_x$, and output tokens from a corresponding \textbf{output vocabulary} $\vocab_y$.

\paragraph{Neural encoder--decoders}
Our approach builds on the standard neural encoder--decoder model with attention \citep{bahdanau2014neural}. In this model, an \textbf{encoder}
represents the input sequence $[x_1, \ldots, x_n]$ as a sequence of representations $[e_1, \ldots, e_n]$
\begingroup
\begin{equation}
    e = \texttt{encoder}(x) 
\end{equation}
\endgroup
Next, a \textbf{decoder} generates a distribution over output sequences $y$ according to the sequentially:
\begingroup
\begin{equation}
    \log p(y \mid x) = \sum\limits_{i=1}^{y} \log p(y_i \mid y_{<i}, e, x) 
\end{equation}
\endgroup
Here we specifically consider decoders with \textbf{attention}.\footnote{All experiments in this paper use LSTM encoders and decoders, but it could be easily integrated with CNNs or transformers (\citealt{Gehring2017ConvolutionalST}; \citealt{Vaswani2017AttentionIA}). We only assume access to a final layer $h_i$, and final attention weights $\alpha_i$; their implementation does not matter.}
When predicting each output token $y_i$, we assign each input token an \textbf{attention weight} $\alpha_i^j$ as in \cref{eq:alphas}. Then, we construct a context representation $c_i$ as the weighted sum of encoder representations $e_i$:
\begin{align}
    \alpha_{i}^{j} &\propto  \exp({h_i^{\top} \, W_{att} \, e_j}) \label{eq:alphas}\\
    c_i &= \sum\limits_{j=1}^{|x|} \alpha_{i}^{j} e_j \label{eq:ci} 
\end{align}
The output distribution over $\vocab_y$, which we denote $p_{\textrm{write},i}$, is calculated by a final projection layer:
\begin{align}
\label{eq:pwrite}
    \hspace{-.75em}
    p(y_i\! =\! w|x) = p_{\textrm{write}_{i}}(w) &\propto \exp(W_\textrm{write}[c_i,h_i])
\end{align}

\paragraph{Copying}

A popular extension of the model described above is the \textbf{copy mechanism}, in which output tokens can be copied from the input sequence in addition to being generated directly by the decoder \citep{jia2016data, see2017get}.
Using the decoder hidden state $h_i$ from above, the model first computes a \textbf{gate probability}:
\begin{equation}
    p_\text{gate} = \sigma(w_\text{gate}^\top h_i)
\end{equation}
and then uses this probability to interpolate between the distribution in \cref{eq:pwrite} and a \textbf{copy distribution} that assigns to each word in the output vocabulary a probability proportional to that word's weight in the attention vector over the input:
\begin{align}
\hspace{-.5em}
    p_\text{copy}(y_i = w \mid x) &= \sum\limits_{j=1}^{|x|} \mathds{1}[{x_j = w}] \cdot \alpha_i^j \\
    p(y_i = w \mid x) &= p_\text{gate} \cdot p_\text{write} (y_i = w \mid x) \nonumber \\
    & \hspace{-2em} + (1 - p_\text{gate}) \cdot p_\text{copy}(y_i = w \mid x)
\end{align}
(note that this implies $\vocab_y \supseteq \vocab_x$).

Content-independent copying is particularly useful in tasks like summarization and machine translation where rare words (like names) are often reused between the input and output.

\begin{CJK*}{UTF8}{bsmi}
\begin{figure}[t!]
    \centering
    \includegraphics[width=0.4\textwidth]{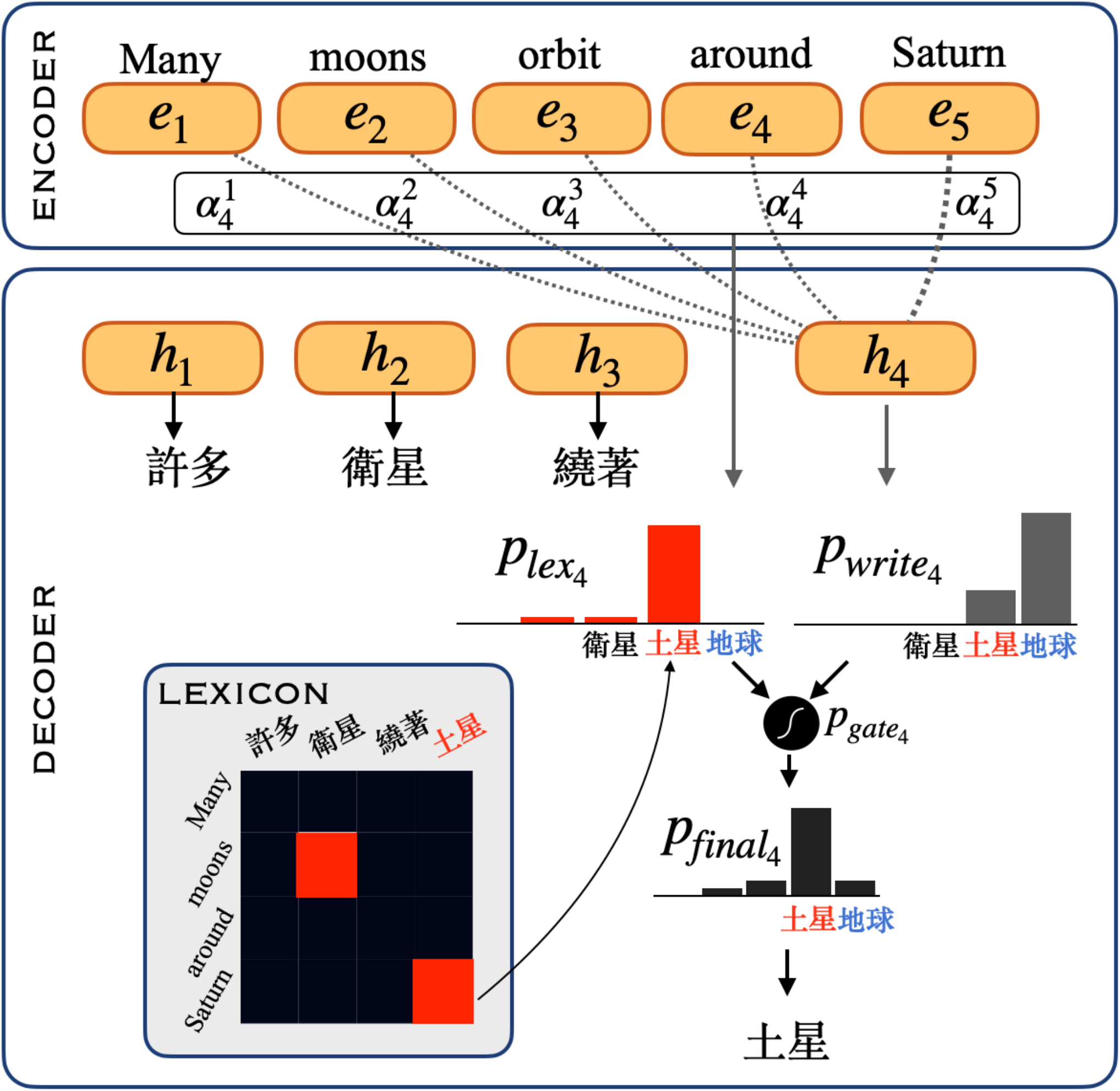}
    \caption{An encoder-decoder model with a lexical translation mechanism applied to English-to-Chinese translation. At decoder step $t=4$,
    attention is focused on the English token \emph{Saturn}. The lexical
    translation mechanism is activated by $p_\text{gate}$, and the model outputs
    the token  \textcolor{red}{土星} directly from the lexicon.
    \textcolor{blue}{地球} means \emph{Earth} and appears much more frequently
    than \emph{Saturn} in the training set.} %
    
    \label{fig:model}
\end{figure}
\end{CJK*}

\paragraph{Our model: Lexical translation} When the input and output
vocabularies are significantly different, copy mechanisms cannot provide further
improvements on a sequence-to-sequence model. However, even for disjoint
vocabularies as in \cref{tab:Colors}, there may be strict correspondences
between individual words on input and output vocabularies, e.g.\ $\texttt{zup}
\mapsto \YELC$ in \cref{tab:Colors}. Following this intuition, the
\textbf{lexical translation mechanism} we introduce in this work extends the
copy mechanism by introducing an additional layer of indirection between the input sequence $x$ and the output prediction $y_i$ as shown in \cref{fig:model}. Specifically, after selecting an input token $x_j \in \vocab_x$, the decoder can ``translate'' it to a context-independent output token $\in \vocab_y$ prior to the final prediction.
We equip the model with an additional \textbf{lexicon parameter} L, a $|\vocab_x| \times |\vocab_y|$ matrix 
in which $\sum_w L_{vw} = 1$, and finally define
\begin{align}
    p_\text{lex}(y_i = w \mid x) &= \sum\limits_{j=1}^{|x|} L_{x_jw} \cdot \alpha_i^j \\
    p(y_i = w \mid x) &= p_\text{gate} \cdot p_\text{write} (y_i = w \mid x) \nonumber \\
    & \hspace{-2em} + (1 - p_\text{gate}) \cdot p_\text{lex}(y_i = w \mid x)
\end{align}
The model is visualized in \cref{fig:model}.
Note that when $\vocab_x = \vocab_y$ and $L = I$ is diagonal, this is identical to the original copy mechanism.
However, this approach can in general be used to produce a larger set of tokens. As shown in \cref{tab:datasets}, coherent token-level translation rules can be identified for many tasks; the lexical translation mechanism allows them to be stored explicitly, using parameters of the base sequence-to-sequence model to record general structural behavior and more complex, context-dependent translation rules.

\section{Initializing the Lexicon}

The lexicon parameter $L$ in the preceding section can be viewed as an ordinary fully-connected layer inside the copy mechanism, and trained end-to-end with the rest of the network. As with other neural network parameters, however, our experiments will show that the initialization of the parameter $L$ significantly impacts downstream model performance, and specifically benefits from initialization with a set of input--output mappings learned with an offline lexicon learning step. Indeed, while not widely used in neural sequence models (though c.f.\ \cref{sec:related}), lexicon-based initialization was a standard feature of many complex non-neural sequence transduction models, including semantic parsers \citep{Kwiatkowski2011LexicalGI} and phrase-based machine translation systems \citep{Koehn2003StatisticalPT}.

But an important distinction between our approach and these others is the fact that we can handle outputs that are not (transparently) compositional. Not every fragment of an input will correspond to a fragment of an output: for example, \emph{thrice} in SCAN has no corresponding output token and instead describes a structural transformation. Moreover, the lexicon is not the only way to generate: complex mappings can also be learned by $p_\text{write}$ without going through the lexicon at all. 

Thus, while most existing work on lexicon learning aims for complete coverage of all word meanings, the model described in \cref{sec:approach-lex} benefits from a lexicon with \emph{high-precision} coverage of \emph{rare phenomena} that will be hard to learn in a normal neural model. Lexicon learning is widely studied in language processing and cognitive modeling, and several approaches with very different inductive biases exist. To determine how to best initialize $L$, we begin by reviewing three algorithms in \cref{sec:other-lex}, and identify ways in which each of them fail to satisfy the high precision criterion above. In \cref{sec:deterministic-lex}, we introduce a simple new lexicon learning rule that addresses this shortcoming.

\subsection{Existing Approaches to Lexicon Learning}
\label{sec:other-lex}

\paragraph{Statistical alignment}\label{ibmm2}

In the natural language processing literature, the IBM translation models
\citep{brown1993mathematics} have served as some of the most popular procedures
for learning token-level input--output mappings. While originally developed for
machine translation, they have also been used to initialize semantic lexicons for semantic parsing \citep{Kwiatkowski2011LexicalGI} and grapheme-to-phoneme conversion \citep{rama2009modeling}. We initialize the lexicon parameter $L$ using Model 2.

Model 2 defines a generative process in which source words $y_i$ are generated from target words $x_j$ via latent alignments $a_i$. Specifically, given a (source, target) pair with $n$ source words and $m$ target words, the probability that the target word $i$ is aligned to the source word $j$ is:
\begingroup
\begin{equation}
    p(a_i = j) \propto \exp\big( -\Big| \frac{i}{m} - \frac{j}{n} \Big| \big)
\end{equation}
\endgroup
Finally, each target word is generated by its aligned source word via a parameter $\theta$:
$
    p(y_i = w) = \theta(v, x_{a_i})
$.
Alignments $a_i$ and lexical parameters $\theta$ can be jointly estimated using the expectation--maximization algorithm \citep{dempster1977maximum}.

In neural models, rather than initializing lexical parameters $L$ directly with corresponding IBM model parameters $\theta$, we run Model 2 in both the forward and reverse directions, then extract counts by \emph{intersecting} these alignments and applying a softmax with temperature $\tau$:
\begin{equation}
   L_{vw} \propto \exp\big(\tau^{-1}\sum_{(x,y) 
   } \sum\limits_{i=1}^{|y|} \mathds{1}[x_{a_i} = v] \mathds{1}[y_i= w]\big)
\end{equation}
\vspace{-3pt}

\noindent
For all lexicon methods discussed in this paper, if an input $v$ is not aligned to any output $w$, we map it to itself if  $\vocab_{x} \subseteq \vocab_{y}$. 
Otherwise we align it uniformly to any unmapped output words (a \emph{mutual exclusivity bias}, \citealt{gandhi2020mutual}).

\paragraph{Mutual information}

Another, even simpler procedure for building a lexicon is based on identifying pairs that have high \emph{pointwise mutual information}. We estimate this quantity directly from co-occurrence statistics in the training corpus:
\begin{equation}
    \textrm{pmi}(v ; w) = \log \frac{\textrm{\#}(v, w)}{\textrm{\#}(v)  \textrm{\#}(w)} + \log |D_{train}|
\end{equation}
where $\#(w)$ is the number of times the word $w$ appears in the training corpus and $\#(w, v)$ is the number of times that $w$ appears in the input and $v$ appears in the output. Finally, we populate the parameter $L$ via a softmax transformation:
$
   L_{vw} \propto \exp((1/\tau) ~\textrm{pmi}\left(v ; w\right))
$.

\paragraph{Bayesian lexicon  learning}
Last, we explore the Bayesian cognitive model of lexicon learning described by \citet{Frank2007ABF}. %
Like IBM model 2, this model is defined by a generative process; here, however, the lexicon itself is part of the generative model. A lexicon $\ell$ is an (unweighted, many-to-many) map defined by a collection of pairs (x, y) with a \emph{description length} prior:
    $p(\ell) \propto e^{-|\ell|}$
(where $|\ell|$ is the number of (input, output) pairs in the lexicon). As in Model 2, given a meaning $y$ and a natural-language description $x$, each $x_i$ is generated independently. We define the probability of a word being used \emph{non-referentially} as
    $p_\text{NR}(x_i \mid \ell) \propto 
    1$ if $x_i \not\in \ell$ and 
    $\kappa$ otherwise.
The probability of being used referentially is:
    $p_\text{R}(x_j \mid y_i, \ell) \propto \mathds{1}_{(x_j, y_i) \in \ell}$.
Finally,
\begin{align}
    p(x_j \mid y_i, \ell) &= (1 - \gamma) p_\text{NR}(x_j \mid \ell) \nonumber \\ &\quad + \gamma \sum\limits_{i=1}^{|y|} p_\text{R}(x_j \mid y_i, \ell)
\end{align}
To produce a final lexical translation matrix $L$ for use in our experiments, we set
    $L_{vw} \propto \exp((1/\tau) ~ p((v, w) \in \ell))$: each entry in $L$ is
    the posterior probability that the given entry appears in a lexicon under
    the generative model above. Parameters are estimated using the Metropolis--Hastings algorithm, with details described in \cref{sec:hyperparameters}. %

\subsection{A Simpler Lexicon Learning Rule} 
\label{sec:deterministic-lex}

\begin{figure}[t]
    \centering
    \includegraphics[width=0.4\textwidth]{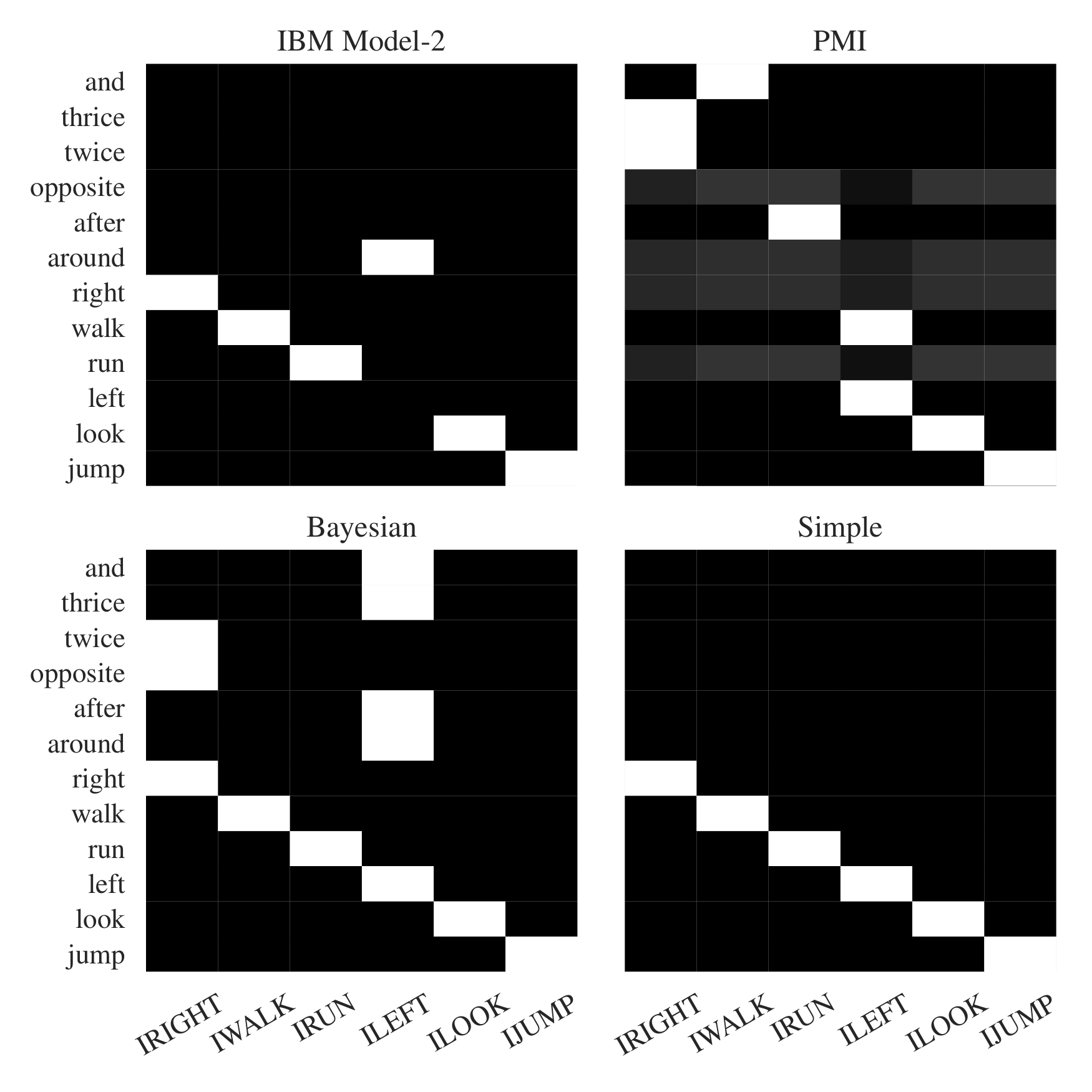}
    \vspace{-.5em}
    \caption{Learned lexicons for the \emph{around right} split in SCAN
    ($\tau=0.1$). The rule-based lexicon learning procedure (\emph{Simple})
    produces correct alignments, while other methods fail due to the correlation
    between \emph{around} and \emph{left} in training data. }
    \label{fig:lexicon_table}
\end{figure}

Example lexicons learned by the three models above are depicted in
\cref{fig:lexicon_table} for the SCAN task shown in \cref{tab:datasets}.
Lexicons learned for remaining tasks can be found in
\cref{sec:learned_lexicons}. It can be seen that all three models produce
errors: the PMI and Bayesian lexicons contain too many entries (in both cases,
numbers are associated with the \texttt{turn right} action and prepositions are
associated with the \texttt{turn left} action). For the IBM model, one of the
alignments is confident but wrong, because the \emph{around} preposition is
associated with \texttt{turn left} action.
In order to understand these errors, and to better characterize the difference
between the demands of lexical translation model initializers and past lexicon
learning schemes,
we explore a simple logical procedure for
extracting lexicon entries that, surprisingly, matchers or outperforms all three
baseline methods in most of our experiments.

What makes an effective, precise lexicon learning rule?
As a first step, consider a maximally restrictive criterion (which we'll call $C_1$) that extracts only pairs $(v, w)$ for which the presence of $v$ in the input is a \emph{necessary and sufficient condition} for the presence of $w$ in the output.
\begin{align}
 &\textit{nec.}(v, w)=\forall xy. ~(w \in y) \to (v \in x) \\
 &\textit{suff.}(v, w)=\forall xy. ~ (v \in x) \to (w \in y) \\
 &C_1(v,w) = \textit{nec.}(v, w) \land \textit{suff.}(v, w)
\end{align}

\noindent
$C_1$ is too restrictive:
in many language understanding problems,
the mapping from \emph{surface forms} to meanings is many-to-one (in \cref{tab:datasets}, both \emph{blessed} and \emph{bless} are associated with the logical form \texttt{bless}). Such mappings cannot be learned by the algorithm described above. We can relax the necessity condition slightly, 
requiring \emph{either} that $v$ is a necessary condition for $w$, or is part of a group
that collectively explains all occurrences of $w$:
\begingroup
\setlength\abovedisplayskip{4pt}
\setlength\belowdisplayskip{4pt}
\begin{align}
    &\textit{no-winner}(w) = \nexists v'. ~ C_1(v',w) \\[0.5em]
    &C_2(v, w) = \textit{suff.}(v, w) ~ \land \nonumber \\ 
    &\hspace{4.8em} \left(\textit{nec.}(v, w) \lor \textit{no-win.}(w)\right) 
\end{align}
\endgroup

As a final refinement,
we note that $C_2$ is likely to capture function words that are present in most sentences, and exclude these by restricting the lexicon to words below a certain frequency threshold:
\begingroup
\begin{equation}
    C_3 = C_2 \, \land \, \left|\{v': \textit{suff.}(v',w)\}\right| \leq \epsilon
\end{equation}
\endgroup
\begin{table*}[t]
\centering
\footnotesize
\begin{tabular}{@{}lcccc@{}}
\toprule
 &  \multicolumn{1}{c}{Colors}  & \multicolumn{1}{c}{jump (SCAN)} & \multicolumn{1}{c}{around right (SCAN)} &  \multicolumn{1}{c}{COGS} \\ \midrule
 LSTM                  & 0.00 \stderr{0.00}     & 0.00 \stderr{0.00}     & 0.09 \stderr{0.05}     & 0.51 \stderr{0.05}       \\
GECA                  & 0.41 \stderr{0.11}     & \bf 1.00 \stderr{0.00} & \bf 0.98 \stderr{0.02} & 0.48 \stderr{0.05}       \\
SyntAtt               & 0.57 \stderr{0.26}     & 0.57 \stderr{0.38}     & 0.28 \stderr{0.26}     & 0.15 \stderr{0.14}       \\
\midrule
LSTM + copy           & -                      & -                      & -                      & 0.66 \stderr{0.03}       \\ 
LSTM + Lex.: Simple   & \bf 0.79 \stderr{0.02} & 0.92 \stderr{0.17}     & 0.95 \stderr{0.01}     & \bf 0.82 \stderr{0.01}   \\
LSTM + Lex.: PMI      & 0.41 \stderr{0.19}     & 0.95 \stderr{0.08}     & 0.02 \stderr{0.04}     & \bf 0.82 \stderr{0.00}   \\
LSTM + Lex.: IBMM2    & \bf 0.79 \stderr{0.02} & 0.79 \stderr{0.27}     & 0.00 \stderr{0.00}     & \bf 0.82 \stderr{0.00}   \\
LSTM + Lex.: Bayesian & 0.51 \stderr{0.21}     & 0.82 \stderr{0.21}     & 0.02 \stderr{0.04}     & 0.70 \stderr{0.04}       \\ 
\bottomrule
\end{tabular}%
\caption{Exact match accuracy results for baselines and lexicon learning models on 4 different compositional generalization splits. Errors are standard deviation among 16 different seeds for Colors, 10 seeds for COGS and SCAN. Unbolded numbers are significantly($p<0.01$) worse than the best result in the column. Models with lexical translation mechanisms and Simple initialization consistently improve over ordinary LSTMs.}
\label{tab:my-table}
\end{table*}

\noindent
The lexicon matrix $L$ is computed by taking the word co-occurrence matrix,
zeroing out all entries where $C_3$ does not hold, then computing a softmax:
$
    L_{vw} \propto C_3(v, w) \exp((1/\tau) ~ \#(v,w))
$.
Surprisingly, as shown in \cref{fig:lexicon_table} and and evaluated below, this
rule (which we label Simple) produces the most effective lexicon initializer for
three of the four tasks we study. The simplicity (and extreme conservativity) of
this rule highlight the different demands on $L$ made by our model and more
conventional (e.g.\ machine translation) approaches: the lexical translation
mechanism benefits from a small number of precise mappings rather than a large
number of noisy ones.

\section{Experiments}
\label{sec:experiments}
We investigate the effectiveness of the lexical translation mechanism on
sequence-to-sequence models for four tasks, three focused on compositional
generalization and one on low-resource machine translation. In all experiments,
we use an LSTM encoder--decoder with attention as the base predictor. We compare
our approach (and variants) with two other baselines: GECA (\citealt{geca}; a data
augmentation scheme) and SynAtt (\citealt{russin2019compositional}; an alternative
seq2seq model parameterization). %
Hyper-parameter selection details are given in the \cref{sec:hyperparameters}. Unless otherwise stated, we use $\tau=0$ and do not fine-tune $L$ after initialization.

\subsection{Colors}
\paragraph{Task} The Colors sequence translation task (see \cref{sec:colors_full_results} for full dataset) was developed to measure \emph{human} inductive biases in sequence-to-sequence learning problems. 
It poses an extreme test of low-resource learning for neural sequence models: it has only 14 training examples that combine four named colors and three composition operations that perform concatenation, repetition and wrapping. \citet{liu2020compositional} solve this dataset with a symbolic stack machine; to the best of our knowledge, our approach is the first ``pure'' neural sequence model to obtain non-trivial accuracy.

\paragraph{Results} Both the Simple and IBMM2 initializers produce a lexicon that maps only color words to colors. Both, combined with the lexical translation mechanism, obtain an average test accuracy of 79\% across 16 runs, nearly matching the human accuracy of 81\% reported by \citet{lake2019human}. The two test examples most frequently predicted incorrectly require generalization to longer sequences than seen during training.
More details (including example-level model and human accuracies) are presented in the appendix \cref{sec:colors_full_results}). These results show that LSTMs are quite effective at learning systematic sequence transformation rules from $\approx3$ examples per function word when equipped with lexical translations.  Generalization to longer sequences remains as an important challenge for future work.

\subsection{SCAN}
\paragraph{Task} SCAN \cite{lake2018generalization} is a larger collection of tests of systematic generalization that pair synthetic English commands (e.g.\ \textit{turn left twice and jump}) to action sequences (e.g.\ \texttt{LTURN LTURN IJUMP}) as shown in \cref{tab:datasets}.  Following previous work, we focus on the \emph{jump} and \emph{around right} splits,
each of which features roughly 15,000 training examples, and
evaluate models' ability to perform 1-shot learning of new primitives (\emph{jump}) and zero-shot interpretation of composition rules (\emph{around right}). While these tasks are now solved by a number of specialized approaches, they remain a challenge for conventional neural sequence models, and an important benchmark for new models. %

\paragraph{Results} In the \emph{jump} split, all initializers improve significantly over the base LSTM when combined with lexical translation. %
Most methods achieve 99\% accuracy at least once across seeds. 
These results are slightly behind GECA (in which all runs succeed) but ahead of SynAtt.\footnote{SynAtt results here are lower than reported in the original paper, which discarded runs with a test accuracy of 0\%.}
Again, they show that lexicon learning is effective for systematic generalization, and that simple initializers (PMI and Simple) outperform complex ones.

\subsection{COGS}
\begin{table}[t]
\centering
\resizebox{\columnwidth}{!}{%
\footnotesize
\begin{tabular}{lp{12mm}p{12mm}p{12mm}}
\toprule
Categories & \hfil LSTM &      \hfil   + copy  & \hfil  + simple \\
\midrule
primitive $\rightarrow$ \{subj, obj, inf\}&   \hfil    \mybar{ 2.8}{1.5} & \hfil  \mybar{ 7.464}{0.069 } & \hfil    \mybar{ 7.42}{0.12 } \\
active $\rightarrow$ passive   &\hfil     \mybar{ 7.74}{0.37 } &      \hfil  \mybar{ 4.5}{3.2 } & \hfil  \mybar{ 7.9968}{0.0053 } \\
obj PP $\rightarrow$ subj PP   &    \hfil    \mybar{ 0.00}{0 } &       \hfil  \mybar{ 0.00}{0 } &   \hfil       \mybar{ 0.00}{0 } \\
passive $\rightarrow$ active   &   \hfil    \mybar{1.8}{1.6 } &      \hfil \mybar{ 3.7}{2.2 } & \hfil  \mybar{ 7.9896}{0.0095 } \\
recursion         &  \hfil \mybar{ 0.166}{0.050 } &  \hfil \mybar{ 0.007}{0.011 } &    \hfil \mybar{ 0.054}{0.065 } \\
unacc $\rightarrow$ transitive &     \hfil  \mybar{ 1.2}{2.3 } & \hfil  \mybar{ 0.004}{0.012 } &  \hfil \mybar{ 7.9928}{0.0091 } \\
obj $\rightarrow$ subj proper  &      \hfil \mybar{ 4.4}{3.5 } &  \hfil \mybar{ 7.958}{0.014 } &    \hfil \mybar{ 7.972}{0.026 } \\
subj $\rightarrow$ obj common  &     \hfil  \mybar{ 6.2}{1.3 } & \hfil  \mybar{ 7.926}{0.035 } &    \hfil \mybar{ 7.945}{0.023 } \\ 
PP dative $\leftrightarrow$ obj dative &   \hfil \mybar{ 7.798}{0.080 } &       \hfil \mybar{ 6.0}{1.4 } &      \hfil \mybar{ 7.942}{0.033 } \\
\midrule
all                 &    \hfil \mybar{ 4.08}{0.44 } &    \hfil  \mybar{ 5.27}{0.24 } &   \hfil   \mybar{ 6.567}{0.043 } \\
\bottomrule
\end{tabular}

}
\caption{COGS accuracy breakdown according to syntactic generalization types for word usages. The label $a \rightarrow b$ indicates that syntactic context $a$ appears in the training set and $b$ in the test set.}
    \label{tab:cogs_detailed}
\end{table}
\paragraph{Task} COGS (Compositional Generalization for Semantic Parsing; \citealt{kim2020cogs}) is an automatically generated English-language semantic parsing dataset that tests systematic generalization in learning language-to-logical-form mappings. It includes 24155 training examples. Compared to the Colors and SCAN datasets, it has a larger vocabulary (876 tokens) and finer-grained inventory of syntactic generalization tests (\cref{tab:cogs_detailed}).

\paragraph{Results} 
Notably, because some tokens appear in both inputs and logical forms in the COGS task, even a standard sequence-to-sequence model with copying significantly outperforms the baseline models in the original work of \citet{kim2020cogs}, solving most tests of generalization over syntactic roles for nouns (but performing worse at generalizations over verbs, including passive and dative alternations).
As above, the lexical translation mechanism (with any of the proposed initializers) provides further improvements, mostly for verbs that baselines model incorrectly (\cref{tab:cogs_detailed}).

\subsection{Machine Translation}

\paragraph{Task} 
To demonstrate that this approach is useful beyond synthetic tests of generalization, we evaluate it on a low-resource English--Chinese translation task
(the Tatoeba\footnote{\url{https://tatoeba.org/}} dataset processed by \citealt{kelly_2021}). For our experiments, we split the data randomly into 19222 training and 2402 test pairs. 
\paragraph{Results} Results are shown in \cref{tab:translation}. Models with a lexical translation mechanism obtain modest improvements (up to 1.5 BLEU) over the baseline. Notably, if we restrict evaluation to test sentences featuring English words that appeared only once in the training set, BLEU improves by more than 2 points, demonstrating that this approach is particularly effective at one-shot word learning (or \emph{fast mapping}; \citealt{carey1978acquiring}).
\cref{fig:model} shows an example from this dataset, in which the model learns to reliably translate \emph{Saturn} from a single training example.
GECA, which makes specific generative assumptions about data distributions, does
not generalize to a more realistic low resource MT problem. However,
the lexical translation mechanism remains effective in natural tasks with large vocabularies and complex grammars.

\begin{table}[t]
\centering
\footnotesize
\begin{tabular}{@{}lll@{}}
\toprule
 & \multicolumn{2}{c}{ENG-CHN} \\ \midrule
 &\multicolumn{1}{c}{full} & \multicolumn{1}{c}{1-shot} \\
LSTM	& 24.18 \stderr{0.37}  & 17.47  \stderr{0.64} \\
LSTM + GECA & 23.90 \stderr{0.55}	& 17.94  \stderr{0.43} \\
LSTM + Lex.: PMI	& 24.36 \stderr{0.09}  & 18.46  \stderr{0.13} \\
LSTM + Lex.: Simple	& 24.35 \stderr{0.09}  & 18.46  \stderr{0.19} \\
LSTM + Lex.: IBMM2	& \bf 25.49 \stderr{0.42}  & \bf19.62 \stderr{0.64} \\
\bottomrule
\end{tabular}%
\caption{BLEU scores for English-Chinese translation. \emph{full} shows results on the full test set, and \emph{1-shot} shows results for text examples in which the English text contains a token seen only once during training.
}
\label{tab:translation}
\end{table}

\begin{table}[t]
\centering
\footnotesize
\begin{tabular}{@{}ll@{}}
\toprule
 & \multicolumn{1}{c}{COGS} \\ \midrule
LSTM &  0.51 \stderr{0.06}\\
\boxRight Lex.: Uniform & 0.56 \stderr{0.07}\\
\boxRight Lex.: Simple & 0.82 \stderr{0.01}\\
\boxSpace\boxRight Soft	& \bf 0.83 \stderr{0.00}\\
\boxSpace\boxSpace\boxRight  Learned	& \bf 0.83 \stderr{0.01}\\
\bottomrule
\end{tabular}%
\caption{Ablation experiments on the COGS dataset. \emph{Uniform} shows results for a lexicon initialized to a uniform distribution. \emph{Soft} sets $\tau=0.1$ with the Simple lexicon learning rule (rather than $0$ in previous experiments). \emph{Learned} shows results for a soft lexicon fine-tuned during training. Soft lexicons with or without learning improve significantly ($p<0.01$) but very slightly over fixed initialization.}
\label{tab:ablation}
\end{table}

\subsection{Fine-Tuning the Lexicon}

In all the experiments above, the lexicon was discretized ($\tau = 0$) and frozen prior to training. In this final section, we revisit that decision, evaluating whether the parameter $L$ can be learned from scratch, or effectively fine-tuned along with decoder parameters. Experiments in this section focus on the COGS dataset.

\textit{Offline initialization of the lexicon is crucial.}
Rather than initializing $L$ using any of the algorithms described in
\cref{sec:approach-lex}, we initialized $L$ to a uniform distribution for each
word and optimized it during training. This improves over the base LSTM (\emph{Uniform} in \cref{tab:ablation}), but performs significantly worse than pre-learned lexicons. %

\textit{Benefits from fine-tuning are minimal.} 
We first increased the temperature parameter $\tau$ to 0.1 (providing a ``soft'' lexicon); this gave a 1\% improvement on COGS (\cref{tab:ablation}. \emph{Soft}). Finally, we updated this soft initialization via gradient descent; this provided no further improvement (\cref{tab:ablation}, \emph{Learned}). One important feature of COGS (and other tests of compositional generalization) is perfect \emph{training} accuracy is easily achieved; thus, there is little pressure on models to learn generalizable lexicons. This pressure must instead come from inductive bias in the initializer.

\section{Conclusion}
We have described a \emph{lexical translation mechanism} for representing token-level translation rules in neural sequence models. We have additionally described a simple initialization scheme for this lexicon that outperforms a variety of existing algorithms. Together, lexical translation and proper initialization enable neural sequence models to solve a diverse set of tasks---including semantic parsing and machine translation---that require 1-shot word learning and 0-shot compositional generalization. 
Future work might focus on generalization to longer sequences, learning of atomic but non-concatenative translation rules, and online lexicon learning in situated contexts.

\section*{Acknowledgements}

This work was supported by the MachineLearningApplications initiative at MIT CSAIL and the MIT--IBM Watson AI lab. Computing resources were provided by a gift from NVIDIA through the NVAIL program and by the Lincoln Laboratory Supercloud.

\bibliographystyle{acl_natbib}
\bibliography{references}
\clearpage
\appendix \label{sec:appendix}
\section{Colors Dataset \& Detailed Results}\label{sec:colors_full_results}
Here we present the full dataset in \cref{tab:full_color} from \citet{lake2019human}, and detailed comparisons of each model with human results in \cref{tab:color_test_results}.
\begin{table}[H]
\resizebox{\columnwidth}{!}{%
\begin{tabular}{@{}llll@{}}
\toprule
\multicolumn{2}{c}{TRAIN} & \multicolumn{2}{c}{TEST}   \\ \midrule
\multicolumn{1}{c}{INPUT} & \multicolumn{1}{c}{OUTPUT} & \multicolumn{1}{c}{INPUT} & \multicolumn{1}{c}{OUTPUT}  \\ \midrule
dax & \REDC & zup fep & \YELC \YELC \YELC\\
lug & \BLUEC & zup kiki dax & \REDC \YELC \\
wif & \GREENC & wif kiki zup & \YELC \GREENC \\
zup & \YELC & zup blicket lug &  \YELC \BLUEC \YELC \\
lug fep & \BLUEC \BLUEC \BLUEC & dax blicket zup & \REDC \YELC \REDC \\
dax fep & \REDC \REDC \REDC & wif kiki zup fep & \YELC \YELC \YELC \GREENC \\
lug blicket wif & \BLUEC \GREENC \BLUEC & zup fep kiki lug  & \BLUEC \YELC \YELC \YELC  \\
wif blicket dax & \GREENC \REDC \GREENC & lug kiki wif blicket zup & \GREENC \YELC \GREENC \BLUEC \\
lug kiki wif & \GREENC \BLUEC &  zup blicket wif kiki dax fep  & \REDC  \REDC  \REDC \YELC \GREENC \YELC \\
dax kiki lug & \BLUEC \REDC &  zup blicket zup kiki zup fep &  \YELC  \YELC  \YELC  \YELC  \YELC  \YELC \\
lug fep kiki wif & \GREENC \BLUEC \BLUEC \BLUEC &\\
wif kiki dax blicket lug & \REDC \BLUEC \REDC \GREENC & \\
lug kiki wif fep & \GREENC \GREENC \GREENC \BLUEC & \\
wif blicket dax kiki lug & \BLUEC \GREENC \REDC \GREENC & \\
\bottomrule
\end{tabular}%
}
\caption{Full Colors dataset with Train and Test examples \cite{lake2019human}}
\label{tab:full_color}
\end{table}
\begin{table}[H]
\resizebox{0.5\textwidth}{!}{%
\begin{tabular}{llllll}
\toprule
Test Examples & Simple/IBM-M2 & Bayesian & GECA &  SyntAtt & Human \\
\midrule
zup fep & 1.0\stderr{0.00} &  0.88\stderr{0.33} &  1.0\stderr{0.00} &  0.7\stderr{0.5} & 0.88 \\
zup kiki dax &  1.0\stderr{0.00} &  0.88\stderr{0.33} &  1.0\stderr{0.00} &  0.7\stderr{0.5} & 0.86 \\
wif kiki zup &   1.0\stderr{0.00} &    0.8\stderr{0.4} &  1.0\stderr{0.00} &  0.8\stderr{0.4}& 0.86 \\
dax blicket zup &   1.0\stderr{0.00} &  0.88\stderr{0.33} &  1.0\stderr{0.00} &  0.8\stderr{0.4} & 0.88 \\
zup blicket lug &  0.94\stderr{0.24} &    0.8\stderr{0.4} &  1.0\stderr{0.00} &  0.8\stderr{0.4} & 0.79 \\
wif kiki zup fep &     1.0\stderr{0.00} &    0.3\stderr{0.5} &  0.0\stderr{0.00} &  0.4\stderr{0.00} 5& 0.85 \\
zup fep kiki lug &       1.0\stderr{0.00} &    0.2\stderr{0.4} &  0.0\stderr{0.00} &  0.8\stderr{0.4} & 0.85 \\
lug kiki wif blicket zup & 1.0\stderr{0.00} &    0.4\stderr{0.5} &  0.0\stderr{0.00} &  0.4\stderr{0.5} & 0.65 \\
zup blicket wif kiki dax fep &    0.0\stderr{0.00} &      0.0\stderr{0.00} &  0.0\stderr{0.00} &    0.0\stderr{0} & 0.70 \\
zup blicket zup kiki zup fep &    0.0\stderr{0.00} &      0.0\stderr{0.00} &  0.0\stderr{0.00} &    0.0\stderr{0.00} & 0.75\\
\bottomrule
\end{tabular}
}
\caption{Colors dataset exact match breakdown for each individual test example. Human results are taken from \cite{lake2019human}Fig2.}
\label{tab:color_test_results}
\end{table}

\section{Learned Lexicons}\label{sec:learned_lexicons}
Here we provide lexicons for each model and dataset (see \cref{fig:model} and \cref{fig:lexicon_table} for remaining datasets). For COGS, we show a representative subset of words.
\begin{figure}[H]
    \centering
    \includegraphics[height=0.25\textheight]{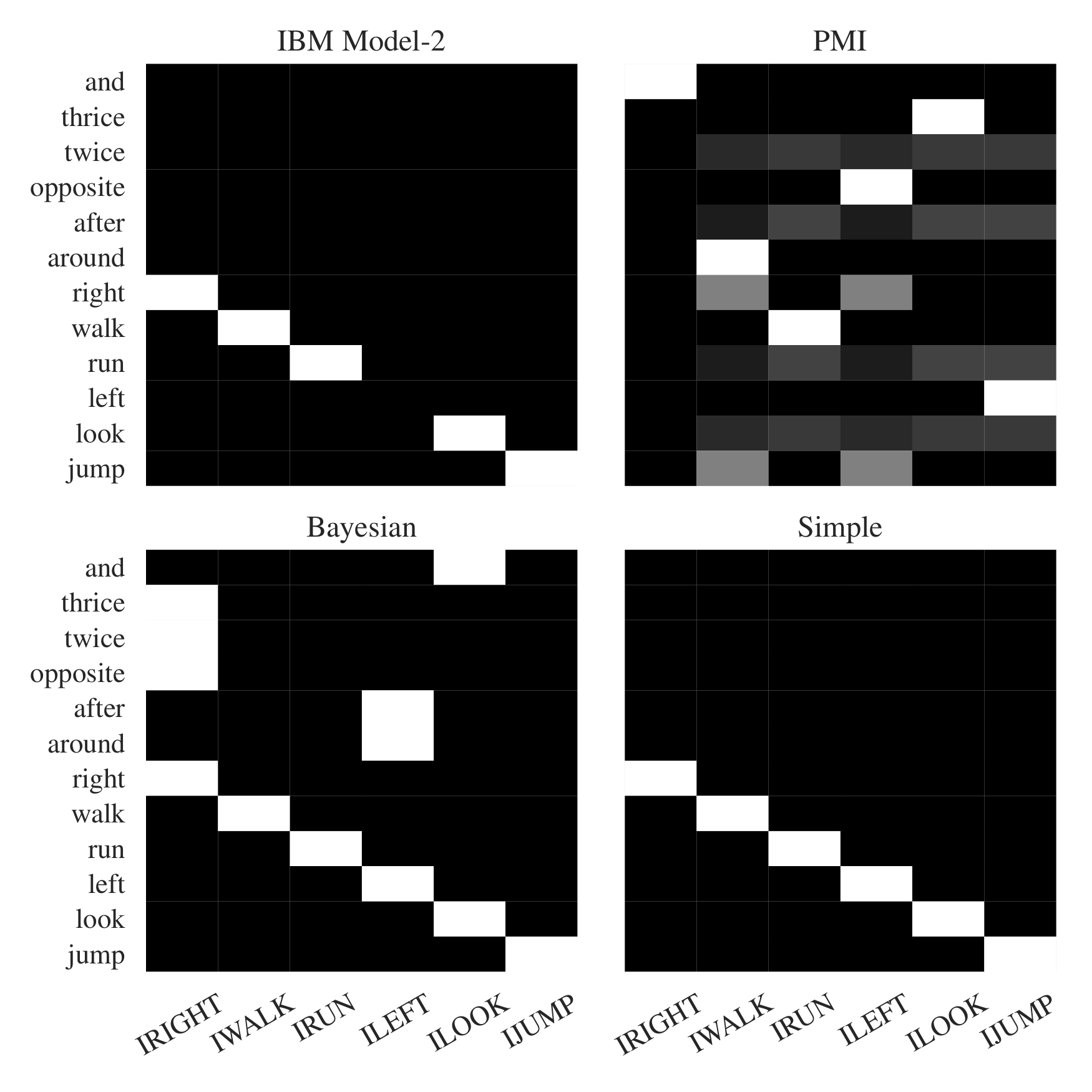}
    \caption{Learned lexicons from SCAN datset \emph{jump} split with $\tau=0.1$}
    \label{fig:jump_lexicon}
\end{figure}

\begin{figure}[H]
    \centering
    \includegraphics[height=0.25\textheight]{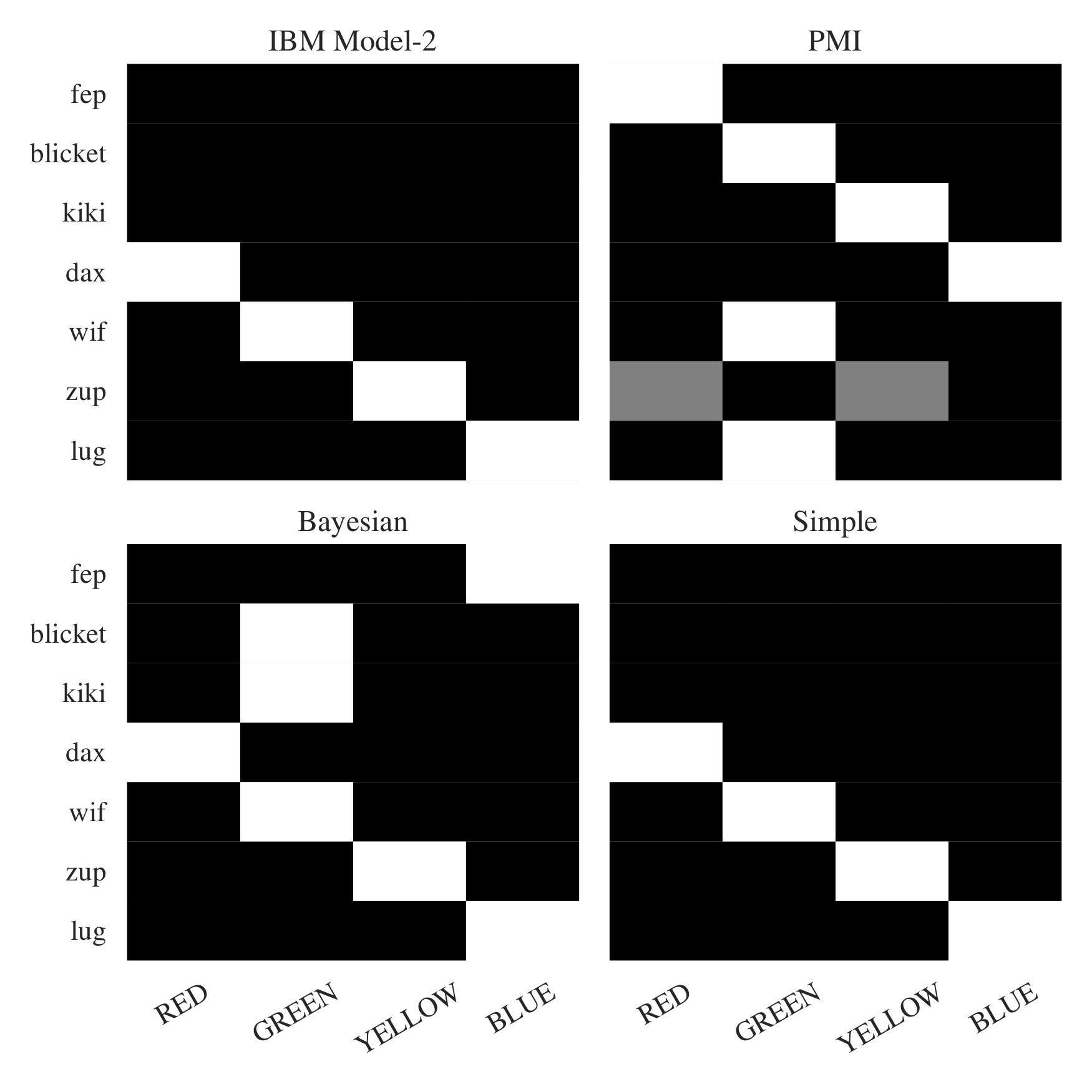}
    \caption{Learned lexicons from Colors datset with $\tau=0.1$}
    \label{fig:colors_lexicon}
\end{figure}

\begin{figure}[H]
    \centering
    \includegraphics[height=0.25\textheight]{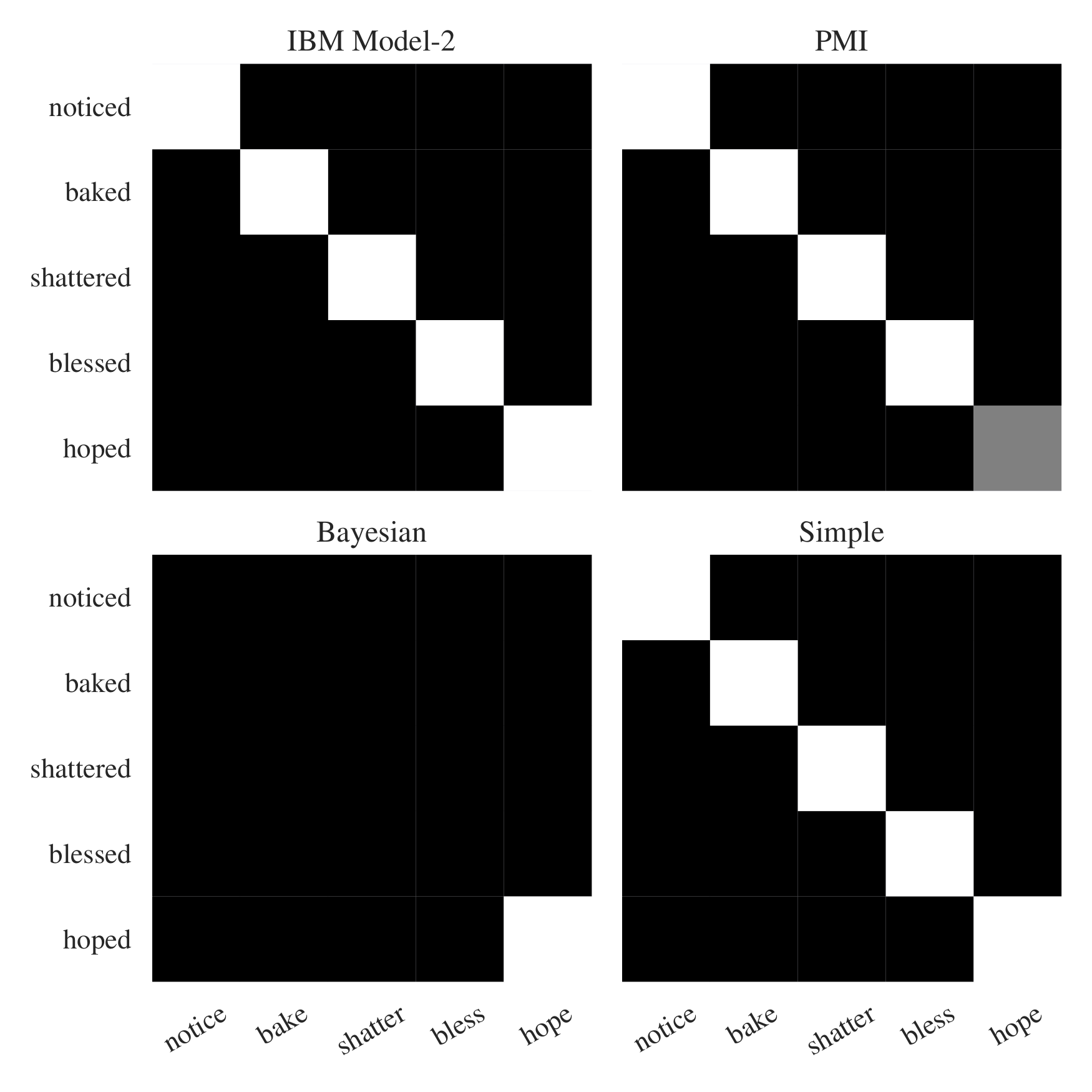}
        \caption{Learned lexicons from COGS datset with $\tau=0.1$. We only show important rare words resposible for our model's improvements over the baseline.}
    \label{fig:cogs_lexicon}
\end{figure}

\section{Hyper-parameter Settings}\label{sec:hyperparameters}
\subsection{Neural Seq2Seq}
Most of the datasets we evaluate do not come with a out-of-distribution validation set, making principled hyperparameter tuning difficult. %
We were unable to reproduce the results of \citet{kim2020cogs} with the hyperparameter settings reported there with our base LSTM setup, and so adjusted them until training was stabilized.
Like the original paper, we used a unidirectional 2-layer LSTM with 512 hidden units, an embedding size of 512, gradient clipping of 5.0, a Noam learning rate scheduler with 4000 warm-up steps, and a batch size of 512. Unlike the original paper, we found it necessary to reduce learning rate to 1.0, increase dropout value to 0.4, and the reduce maximum step size timeout to 8000. 

We use same parameters for all COGS, SCAN, and machine translation experiments. For SCAN and Colors, we applied additional dropout (p=0.5) in the last layer of $p_{\textrm{write}}$.

Since Colors has 14 training examples, we need a different batch size, set to 1/3 of the training set size ($=5$). Qualitative evaluation of gradients in training time revealed that stricter gradient clipping was also needed ($=0.5$). Similarly, we decreased warm-up steps to 32 epochs. All other hyper-parameters remain the same.

\subsection{Lexicon Learning}
\paragraph{Simple Lexicon}
The only parameter in the simple lexicon is $\epsilon$, set to 3 in all experiments.
\paragraph{Bayesian}
The original work of \citet{Frank2007ABF} did not report hyperparemeter settings or sampler details. We found $\alpha = 2$, $\gamma = 0.95$ and $\kappa = 0.1$ to be effective. The M--H proposal distribution inserts or removes a word from the lexicon with 50\% probability. For deletions, an entry is removed uniformly at random. For insertions, an entry is added with probability proportional to the empirical joint co-occurrence probability of the input and output tokens. Results were averaged across 5 runs, with a burn-in period of 1000 and a sample drawn every 10 steps.

\paragraph{IBM Model 2}
We used the FastAlign implementation \cite{dyer2013simple} and experimented with a variety of hyperparameters in the  alignment algorithm itself (favoring diagonal alignment, optimizing tension, using dirichlet priors) and  diagonalization heuristics (grow-diag, grow-diag-final, grow-diag-final-and, union). We found that optimizing tension and using the “intersect” diagonalization heuristic works the best overall.

\section{Baseline Results}
\subsection{GECA}
We reported best results for SCAN dataset from reproduced results in \cite{akyurek2020learning}. For other datasets (COGS and Colors), we performed a hyperparameter search over augmentation ratios of 0.1 and 0.3 and hidden sizes of \{128, 256, 512\}. We report the best results for each dataset.
\subsection{SyntAtt}
We used the public GitHub repository of SyntAtt\footnote{(\url{https://github.com/jlrussin/syntactic_attention})} and reproduced reported results for the SCAN dataset. For other datasets, we also explored ''syntax action'' option, in which both contextualized context (syntax) and un-contextualized embeddings (semantics) used in final layer \citet{russin2019compositional}. We additionally performed a search over hidden layer sizes \{128,256,512\} and depths \{1,2\}. We report the best results for each dataset.

\section{Datasets \& Evaluation \& Tokenization}
\subsection{Datasets and Sizes}
\resizebox{0.49\textwidth}{!}{%
\begin{tabular}{llllll}
\toprule
{} &  around\_right &     jump &   COGS &  Colors &  ENG-CHN \\
\midrule
train &       15225 &  14670 &  24155 &    14 &    19222 \\
validation   &          - &      - &   3000 &     - &     2402 \\
test  &        4476 &   7706 &  21000 &    10 &     2402 \\
\bottomrule
\end{tabular}
}
\subsection{Evaluation}
We report exact match accuracies and BLEU scores. In both evaluations we include punctuation. For BLEU we use NLTK \footnote{\url{https://www.nltk.org/}} library's default implementation.
\subsection{Tokenization}
We use \emph{Moses} library\footnote{\url{https://pypi.org/project/mosestokenizer/}} for English tokenization, and \emph{jieba}\footnote{\url{https://github.com/fxsjy/jieba}} library for Chinese tokenization. In other datasets, we use default space tokenization.
\section{Computing Infrastructure}
Experiments were performed on a DGX-2 with NVIDIA 32GB VOLTA-V100 GPUs. Experiments take at most 2.5 hours on a single GPU.
\end{document}